\documentclass[conference]{IEEEtran}
\ifCLASSINFOpdf
\else
\fi

\usepackage{amsmath}
\usepackage{amsfonts}
\usepackage{booktabs}
\usepackage{graphicx}
\begin{document}
%
\title{Time Series Classification from Scratch with Deep Neural Networks: A Strong Baseline}

\author{\IEEEauthorblockN{Zhiguang Wang, Weizhong Yan}
\IEEEauthorblockA{GE Global Research\\
\{zhiguang.wang, yan\}@ge.com
}
\and
\IEEEauthorblockN{Tim Oates}
\IEEEauthorblockA{Computer Science and Electric Engineering\\
University of Maryland Baltimore County	\\
oates@umbc.edu
}
}


%


\maketitle

\begin{abstract}
We propose a simple but strong baseline for time series classification from scratch with deep neural networks. Our proposed baseline models are pure end-to-end without any heavy preprocessing on the raw data or feature crafting. The proposed Fully Convolutional Network (FCN) achieves premium performance to other state-of-the-art approaches and our exploration of the very deep neural networks with the ResNet structure is also competitive. The global average pooling in our convolutional model enables the exploitation of the Class Activation Map (CAM) to find out the contributing region in the raw data for the specific labels. Our models provides a simple choice for the real world application and a good starting point for the future research. An overall analysis is provided to discuss the generalization capability of our models, learned features, network structures and the classification semantics.
\end{abstract}


%
\IEEEpeerreviewmaketitle

\section{Introduction}
Time series data is ubiquitous. Both human activities and nature produces time series everyday and everywhere, like weather readings, financial recordings, physiological signals and industrial observations. As the simplest type of time series data, univariate time series provides a reasonably good starting point to study such temporal signals. The representation learning and classification research has found many potential application in the fields like finance, industry, and health care. 

However, learning representations and classifying time series are still attracting much attention. As the earliest baseline, distance-based methods work directly on raw time series with some pre-defined similarity measures such as Euclidean distance or Dynamic time warping (DTW) \cite{keogh2005exact} to perform classification. The combination of DTW and the k-nearest-neighbors classifier is known to be a very efficient approach  as a golden standard in the last decade. 

Feature-based methods suppose to extract a set of features that are able to represent the global/local time series patterns. Commonly, these features are quantized to form a Bag-of-Words (BoW), then given to the classifiers \cite{lin2007experiencing}. Feature-based approaches mostly differ in the extracted features. To name a few recent benchmarks, The bag-of-features framework (TSBF) \cite{baydogan2013bag} extracts the interval features with different scales from each interval to form an instance, and each time series forms a bag. A supervised codebook is built with the random forest for classifying the time series. Bag-of-SFA-Symbols (BOSS) \cite{schafer2015boss} proposes a distance based on the histograms of symbolic Fourier approximation words. Its extension, the BOSSVS method \cite{schafer2015scalable} combines the BOSS model with the vector space model to reduce the time complexity and improve the performance by ensembling the models with difference window size. The final classification is performed with the One-Nearest-Neighbor classifier. 

Ensemble based approaches combine different classifiers together to achieve a higher accuracy. Different ensemble paradigms integrate various feature sets or classifiers. The Elastic Ensemble (PROP) \cite{lines2015time} combines 11 classifiers based on elastic distance measures with a weighted ensemble scheme. Shapelet ensemble (SE) \cite{bagnall2015time} produces the classifiers through the shapelet transform in conjunction with a heterogeneous ensemble. The flat collective of transform-based ensembles (COTE) is an ensemble of 35 different classifiers based on the features extracted from both the time and frequency domains.

All the above approaches need heavy crafting on data preprocessing and feature engineering. Recently, some effort has been spent to exploit the deep neural network, especially convolutional neural networks (CNN) for end-to-end time series classification. In \cite{zheng2016exploiting}, a multi-channel CNN (MC-CNN) is proposed for multivariate time series classification. The filters are applied on each single channel and the features are flattened across channels as the input to a fully connected layer. The authors applied sliding windows to enhance the data. They only evaluate this approach on two multivariate time series datasets, where there is no published benchmark for comparison. In \cite{cui2016multi}, the author proposed a multi-scale CNN approach (MCNN) for univariate time series classification. Down sampling, skip sampling and sliding windows are used for preprocessing the data to manually prepare for the multi-scale settings. Although this approach claims the state-of-the-art performance on 44 UCR time series datasets \cite{chen2016ucr}, the heavy preprocessing efforts and a large set of hyperparameters make it complicated to deploy. The proposed window slicing method for data augmentation seems to be ad-hoc. 

\begin{figure*}[t]
	\centering
	\includegraphics[width=0.9\textwidth]{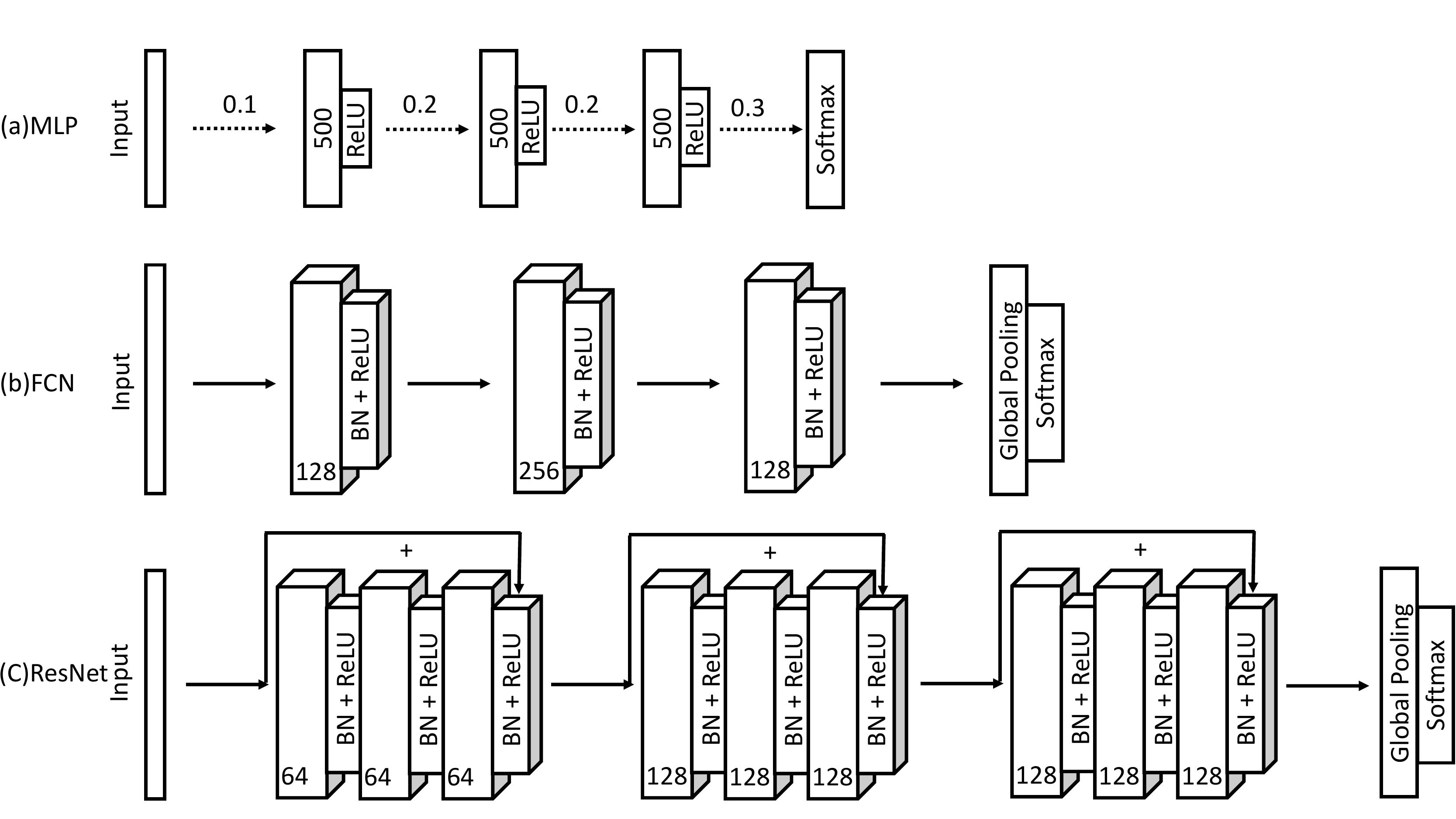}
	\caption{The network structure of three tested neural networks. Dash line indicates the operation of dropout.}
	\label{fig:Archi}
\end{figure*}

We provide a standard baseline to exploit deep neural networks for end-to-end time series classification without any crafting in feature engineering and data preprocessing. The deep multilayer perceptrons (MLP), fully convolutional networks (FCN) and the residual networks (ResNet) are evaluated on the same 44 benchmark datasets with other benchmarks. Through a pure end-to-end training on the raw time series data , the ResNet and FCN achieve comparable or better performance than COTE and MCNN. The global average pooling in our convolutional model enables the exploitation of the Class Activation Map (CAM) to find out the contributing region in the raw data for the specific labels. 

\section{Network Architectures}
We tested three deep neural network architectures to provide a fully comprehensive baseline. 

\subsection{Multilayer Perceptrons}
Our plain baselines are basic MLP by stacking three fully-connected layers. The fully-connected layers each has 500 neurons following two design rules: (i) using dropout \cite{srivastava2014dropout} at each layer's input to improve the generalization capability ; and (ii) the non-linearity is fulfilled by the rectified linear unit (ReLU)\cite{nair2010rectified} as the activation function to prevent saturation of the gradient when the network is deep. The network ends with a softmax layer. A basic layer block is formalized as

\begin{eqnarray}
\tilde{x} = f_{dropout,p}(x) \nonumber \\
y = \bf{W} \cdot \tilde{x} + b \nonumber \\
h = ReLU(y)
\label{eqn:mlp}
\end{eqnarray}

This architecture is mostly distinguished from the seminal MLP decades ago by the utilization of ReLU and dropout. ReLU helps to stack the networks deeper and dropout largely prevent the co-adaption of the neurons to help the model generalizes well especially on some small datasets. However, if the network is too deep, most neuron will hibernate as the ReLU totally halve the negative part. The Leaky ReLU \cite{xu2015empirical} might help, but we only use three layers MLP with the ReLU to provide a fundamental baselines. The dropout rates at the input layer, hidden layers and the softmax layer are \{0.1, 0.2, 0.3\}, respectively (Figure \ref{fig:Archi}(a)).

\subsection{Fully Convolutional Networks}
FCN has shown compelling quality and efficiency for semantic segmentation on images \cite{long2015fully}. Each output pixel is a classifier corresponding to the receptive field and the networks can thus be trained pixel-to-pixel given the category-wise semantic segmentation annotation. 

In our problem settings, the FCN is performed as a feature extractor. Its final output still comes from the softmax layer. The basic block is a convolutional layer followed by a batch normalization layer \cite{ioffe2015batch} and a ReLU activation layer. The convolution operation is fulfilled by three 1-D kernels with the sizes $\{8, 5,3\}$ without striding. The basic convolution block is 

\begin{eqnarray}
y = \bf{W} \otimes x + b \nonumber \\
s = BN(y) \nonumber \\
h = ReLU(s)
\label{eqn:conv}
\end{eqnarray}

$\otimes$ is the convolution operator. We build the final networks by stacking three convolution blocks with the filter sizes \{128, 256, 128\} in each block. Unlike the MCNN and MC-CNN, We exclude any pooling operation. This strategy is also adopted in the ResNet \cite{he2015deep} as to prevent overfitting. Batch normalization is applied to speed up the convergence speed and help improve generalization. After the convolution blocks, the features are fed into a global average pooling layer \cite{lin2013network} instead of a fully connected layer, which largely reduces the number of weights. The final label is produced by a softmax layer (Figure \ref{fig:Archi}(b)). 

\subsection{Residual Network}
ResNet extends the neural networks to a very deep structures by adding the shortcut connection in each residual block to enable the gradient flow directly through the bottom layers. It achieves the state-of-the-art performance in object detection and other vision related tasks \cite{he2015deep}. We explore the ResNet structure since we are really interested to see how the very deep neural networks perform on the time series data. Obviously, the ResNet overfits the training data much easier because the datasets in UCR is comparatively small and lack of enough variants to learn the complex structures with such deep networks, but it is still a good practice to import the much deeper model and analyze the pros and cons. 

We reuse the convolutional blocks in Equation \ref{eqn:conv} to build each residual block. Let $Block_{k}$ denotes the convolutional block with the number of filters $k$, the residual block is formalized as

\begin{eqnarray}
h_1 = Block_{k_1}(x) \nonumber \\
h_2 = Block_{k_2}(h_1) \nonumber \\
h_3 = Block_{k_3}(h_2) \nonumber \\
y = h_3 + x \nonumber \\
\hat{h} = ReLU(y)
\end{eqnarray}

The number of filters $k_i = \{64, 128, 128\}$. The final ResNet stacks three residual blocks and followed by a global average pooling layer and a softmax layer. As this setting simply reuses the structures of the FCN,  certainly there are better structures for the problem, but our given structures are adequate to provide a  qualified demonstration as a baseline (Figure \ref{fig:Archi}(c)).   

\section{Experiments and Results}
\subsection{Experiment Settings}
We test our proposed neural networks on the same subset of the UCR time series repository, which includes 44 distinct time series datasets, to compare with other benchmarks.  All the dataset has been split into training and testing by default. The only preprocessing in our experiment is z-normalization on both training and test split with the mean and standard deviation of the training part for each dataset. The MLP is trained with Adadelta \cite{zeiler2012adadelta} with learning rate 0.1, $\rho=0.95$ and $\epsilon=1e-8$. The FCN and ResNet are trained with Adam \cite{kingma2014adam} with the learning rate 0.001, $\beta_1 = 0.9, \beta_2 = 0.999$ and $\epsilon = 1e-8$. The loss function for all tested model is categorical cross entropy. We choose the best model that achieves the lowest training loss and report its performance on the test set. While this training setting tends to give us a overfitted configuration and most likely to generalize poorly on the test set, we can see that our proposed networks generalize quite well. Unlike other benchmarks, our experiment excludes the hyperparameter tuning and cross validation to provide a most unbiased baseline. Such settings also largely reduce the complexity for training and deploying the deep learning models. \footnote{The codes are available at https://github.com/cauchyturing/ 
	UCR\_Time\_Series\_Classification\_Deep\_Learning\_Baseline \cite{chollet2015keras}.}  

\subsection{Evaluation}
Table \ref{tab:results} shows the results and a comprehensive comparison with eight other best benchmark methods. We report the test error rate from the best model trained with the minimum cross-entropy loss and the number of dataset on which it achieved the best performance. Some literature (like \cite{cui2016multi,schafer2015scalable}) also report the ranks and other ranking-based statistics to evaluate the performance and make the comparison, so we also provide the average rankings.

However, neither the number of best-performed dataset or the ranking based statistics is an unbiased measurement to compare the performance. The number of best-performed dataset focuses on the top performance and is highly skewed. The ranking based statistics is highly sensitive to the model pools. "Better than" as a comparative measurement is also skewed as the input models might arbitrarily changed. All those evaluation measures wipe out the factor of number of classes. 

We propose a simple evaluation measure, Mean Per-Class Error (MPCE) to evaluate the classification performance of the specific models on multiple datasets. For a given model $M = \{m_i\}$, a dataset pool $D=\{d_k\}$ with the number of class label $C =\{c_k\}$ and the corresponding error rate $E=\{e_k\}$, 

\begin{eqnarray}
PCE_k = \frac{e_k}{c_k} \nonumber \\
MPCE_i = \frac{1}{K}\sum{PCE_k}
\end{eqnarray}

$k$ refers to each dataset and $i$ denotes to each model. The intuition behind MPCE is simple: the expected error rate for a single class across all the datasets. By considering the number of classes, MPCE is more robust as a baseline criterion. A paired T-test on PCE identifies if the differences of the MPCE are significant across different models.

\begin{table*}[t]
  \centering
  \caption{Testing error and the mean per-class error (MPCE) on 44 UCR time series dataset}
    \begin{tabular}{rrrrrrrrr|rrr}
    \toprule
    Err Rate & DTW   & COTE  & MCNN  & BOSSVS & PROP  & BOSS  & SE1   & TSBF  & MLP   & FCN   & ResNet \\
    \midrule
    Adiac & 0.396 & 0.233 & 0.231 & 0.302 & 0.353 & 0.22  & 0.373 & 0.245 & 0.248 & \textbf{0.143} & 0.174 \\
    Beef  & 0.367 & \textbf{0.133} & 0.367 & 0.267 & 0.367 & 0.2   & \textbf{0.133} & 0.287 & 0.167 & 0.25  & 0.233 \\
    CBF   & 0.003 & 0.001 & 0.002 & 0.001 & 0.002 & \textbf{0} & 0.01  & 0.009 & 0.14  & \textbf{0} & 0.006 \\
    ChlorineCon & 0.352 & 0.314 & 0.203 & 0.345 & 0.36  & 0.34  & 0.312 & 0.336 & \textbf{0.128} & 0.157 & 0.172 \\
    CinCECGTorso & 0.349 & 0.064 & 0.058 & 0.13  & 0.062 & 0.125 & \textbf{0.021} & 0.262 & 0.158 & 0.187 & 0.229 \\
    Coffee & \textbf{0} & \textbf{0} & 0.036 & 0.036 & \textbf{0} & \textbf{0} & \textbf{0} & 0.004 & \textbf{0} & \textbf{0} & \textbf{0} \\
    CricketX & 0.246 & \textbf{0.154} & 0.182 & 0.346 & 0.203 & 0.259 & 0.297 & 0.278 & 0.431 & 0.185 & 0.179 \\
    CricketY & 0.256 & 0.167 & \textbf{0.154} & 0.328 & 0.156 & 0.208 & 0.326 & 0.259 & 0.405 & 0.208 & 0.195 \\
    CricketZ & 0.246 & \textbf{0.128} & 0.142 & 0.313 & 0.156 & 0.246 & 0.277 & 0.263 & 0.408 & 0.187 & 0.187 \\
    DiatomSizeR & 0.033 & 0.082 & \textbf{0.023} & 0.036 & 0.059 & 0.046 & 0.069 & 0.126 & 0.036 & 0.07  & 0.069 \\
    ECGFiveDays & 0.232 & \textbf{0} & \textbf{0} & \textbf{0} & 0.178 & \textbf{0} & 0.055 & 0.183 & 0.03  & 0.015 & 0.045 \\
    FaceAll & 0.192 & 0.105 & 0.235 & 0.241 & 0.152 & 0.21  & 0.247 & 0.234 & 0.115 & \textbf{0.071} & 0.166 \\
    FaceFour & 0.17  & 0.091 & \textbf{0} & 0.034 & 0.091 & \textbf{0} & 0.034 & 0.051 & 0.17  & 0.068 & 0.068 \\
    FacesUCR & 0.095 & 0.057 & 0.063 & 0.103 & 0.063 & \textbf{0.042} & 0.079 & 0.09  & 0.185 & 0.052 & 0.042 \\
    50words & 0.31  & 0.191 & 0.19  & 0.367 & \textbf{0.18} & 0.301 & 0.288 & 0.209 & 0.288 & 0.321 & 0.273 \\
    fish  & 0.177 & 0.029 & 0.051 & 0.017 & 0.034 & \textbf{0.011} & 0.057 & 0.08  & 0.126 & 0.029 & \textbf{0.011} \\
    GunPoint & 0.093 & 0.007 & \textbf{0} & \textbf{0}     & 0.007 & \textbf{0} & 0.06  & 0.011 & 0.067 & \textbf{0} & 0.007 \\
    Haptics & 0.623 & 0.488 & 0.53  & 0.584 & 0.584 & 0.536 & 0.607 & 0.488 & 0.539 & \textbf{0.449} & 0.495 \\
    InlineSkate & 0.616 & 0.551 & 0.618 & 0.573 & 0.567 & \textbf{0.511} & 0.653 & 0.603 & 0.649 & 0.589 & 0.635 \\
    ItalyPower & 0.05  & 0.036 & \textbf{0.03} & 0.086 & 0.039 & 0.053 & 0.053 & 0.096 & 0.034 & \textbf{0.03} & 0.04 \\
    Lightning2 & 0.131 & 0.164 & 0.164 & 0.262 & 0.115 & 0.148 & \textbf{0.098} & 0.257 & 0.279 & 0.197 & 0.246 \\
    Lightning7 & 0.274 & 0.247 & 0.219 & 0.288 & 0.233 & 0.342 & 0.274 & 0.262 & 0.356 & \textbf{0.137} & 0.164 \\
    MALLAT & 0.066 & 0.036 & 0.057 & 0.064 & 0.05  & 0.058 & 0.092 & 0.037 & 0.064 & \textbf{0.02} & 0.021 \\
    MedicalImages & 0.263 & 0.258 & 0.26  & 0.474 & 0.245 & 0.288 & 0.305 & 0.269 & 0.271 & \textbf{0.208} & 0.228 \\
    MoteStrain & 0.165 & 0.085 & 0.079 & 0.115 & 0.114 & 0.073 & 0.113 & 0.135 & 0.131 & \textbf{0.05} & 0.105 \\
    NonInvThorax1 & 0.21  & 0.093 & 0.064 & 0.169 & 0.178 & 0.161 & 0.174 & 0.138 & 0.058 & \textbf{0.039} & 0.052 \\
    NonInvThorax2 & 0.135 & 0.073 & 0.06  & 0.118 & 0.112 & 0.101 & 0.118 & 0.13  & 0.057 & \textbf{0.045} & 0.049 \\
    OliveOil & 0.167 & 0.1   & 0.133 & 0.133 & 0.133 & 0.1   & 0.133 & \textbf{0.09}  & 0.60  & 0.167 & 0.133 \\
    OSULeaf & 0.409 & 0.145 & 0.271 & 0.074 & 0.194 & \textbf{0.012} & 0.273 & 0.329 & 0.43  & \textbf{0.012} & 0.021 \\
    SonyAIBORobot & 0.275 & 0.146 & 0.23  & 0.265 & 0.293 & 0.321 & 0.238 & 0.175 & 0.273 & 0.032 & \textbf{0.015} \\
    SonyAIBORobotII & 0.169 & 0.076 & 0.07  & 0.188 & 0.124 & 0.098 & 0.066 & 0.196 & 0.161 & \textbf{0.038} & \textbf{0.038} \\
    StarLightCurves & 0.093 & 0.031 & 0.023 & 0.096 & 0.079 & \textbf{0.021} & 0.093 & 0.022 & 0.043 & 0.033 & 0.029 \\
    SwedishLeaf & 0.208 & 0.046 & 0.066 & 0.141 & 0.085 & 0.072 & 0.12  & 0.075 & 0.107 & \textbf{0.034} & 0.042 \\
    Symbols & 0.05  & 0.046 & 0.049 & \textbf{0.029} & 0.049 & 0.032 & 0.083 & 0.034 & 0.147 & 0.038 & 0.128 \\
    SyntheticControl & 0.007 & \textbf{0} & 0.003 & 0.04  & 0.01  & 0.03  & 0.033 & 0.008 & 0.05  & 0.01  &  \textbf{0} \\
    Trace & \textbf{0} & 0.01  & \textbf{0} & \textbf{0} & 0.01  & \textbf{0} & 0.05  & 0.02  & 0.18  & \textbf{0} & \textbf{0} \\
    TwoLeadECG & \textbf{0} & 0.015 & 0.001 & 0.015 & \textbf{0} & 0.004 & 0.029 & 0.001 & 0.147 & \textbf{0} & \textbf{0} \\
    TwoPatterns & 0.096 & \textbf{0} & 0.002 & 0.001 & 0.067 & 0.016 & 0.048 & 0.046 & 0.114 & 0.103 & \textbf{0} \\
    UWaveX & 0.272 & 0.196 & 0.18  & 0.27  & 0.199 & 0.241 & 0.248 & \textbf{0.164} & 0.232 & 0.246 & 0.213 \\
    UWaveY & 0.366 & 0.267 & 0.268 & 0.364 & 0.283 & 0.313 & 0.322 & \textbf{0.249} & 0.297 & 0.275 & 0.332 \\
    UWaveZ & 0.342 & 0.265 & 0.232 & 0.336 & 0.29  & 0.312 & 0.346 & \textbf{0.217} & 0.295 & 0.271 & 0.245 \\
    wafer & 0.02  & \textbf{0.001} & 0.002 & \textbf{0.001} & 0.003 & \textbf{0.001} & 0.002 & 0.004 & 0.004 & 0.003 & 0.003 \\
    WordSynonyms & 0.351 & 0.266 & 0.276 & 0.439 & \textbf{0.226} & 0.345 & 0.357 & 0.302 & 0.406 & 0.42  & 0.368 \\
    yoga  & 0.164 & 0.113 & 0.112 & 0.169 & 0.121 & \textbf{0.081} & 0.159 & 0.149 & 0.145 & 0.155 & 0.142 \\
    \midrule
    Win   & 3     & 8     & 7     & 5     & 4     & 13    & 4     & 4     & 2     & \textbf{18} & 8 \\
	AVG Arithmetic ranking & 8.205 & \textbf{3.682} & 3.932 & 7.318 & 5.545 & 4.614 & 7.455 & 6.614 & 7.909 & 3.977 & 4.386 \\
	AVG geometric ranking & 7.160 & 3.054 & 3.249 & 5.997 & 4.744 & 3.388 & 6.431 & 5.598 & 6.941 & \textbf{2.780} & 3.481 \\
    MPCE  & 0.0397 & 0.0226 & 0.0241 & 0.0330 & 0.0304 & 0.0256 & 0.0302 & 0.0335 & 0.0407 & \textbf{0.0219} & 0.0231 \\
    \bottomrule
    \end{tabular}%
  \label{tab:results}%
\end{table*}%

\subsection{Results and Analysis}
We select seven existing best methods\footnote{'Best' means the overall performance is competitive and the model should achieve the best performance on at least 4 datasets (10\% of the all the 44 datasets).} that claim the state-of-the-art results and published within recent three years: time series based on a bag-offeatures (TSBF), Elastic Ensemble (PROP), 1-NN Bag-Of-SFA-Symbols (BOSS) in Vector Space (BOSSVS), the Shapelet Ensemble (SE1) model, flat-COTE (COTE) and multi-scale CNN (MCNN). Note that COTE is an ensemble model which combines the weighted votes over 35 different classifiers. BOSSVS is an ensemble of multiple BOSS models with different window length. 1NN-DTW is also included as a simple standard baseline. The training and deploying complexity of our models are small like 1NN-DTW as their pipeline is all from scratch without any heavy pre-processing and data augmentations, while our baselines do not need feature crafting.

\begin{figure*}[t]
	\centering
	\includegraphics[width=1.05\textwidth]{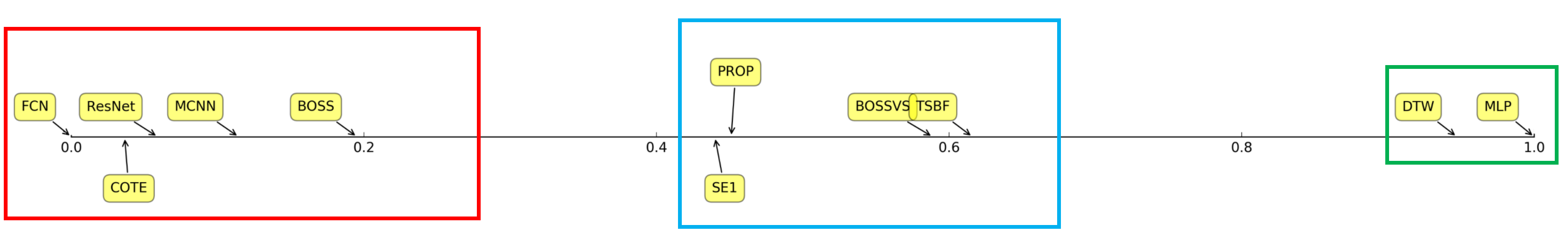}
	\caption{Models grouping by the paired T-test of means on the normalized PCE scores.}
	\label{fig:Ranking}
\end{figure*}

In Table \ref{tab:results}, we provide four metrics to fully evaluate different approaches. FCN indicates the best performance on three metrics at the first sight, while ResNet is also competitive on the MPCE score and rankings.

In \cite{cui2016multi,schafer2015scalable}, the authors proposed to validate the effectiveness of their models by Wilcoxon signed-rank test on the error rates. Instead, we choose the Wilcoxon rank-sum test as it can deal with the tie conditions among the error rates with the tie correction (Appenix Table \ref{tab:Mann–Whitney}). The p-values in our case are quite different with the results reported by \cite{cui2016multi}. Except for MLP and DTW, all other approaches are 'linked' together based on the p-value. It possibly because the model pool we choose are different and the ranking based statistics is very sensitive to the model pool and its size.

The MPCE score is reported in the last row. FCN and MLP have the best and worse MPCE score respectively. The ResNet ranks 3rd among all the 11 models, just a little worse than COTE. A paired T-test of mean on the PCE score is performed to tell if the difference of MPCE is significant (Appendix Table \ref{tab:PCE-test}). Interestingly, we found the difference of MPCE among COTE, MCNN, BOSS, FCN and ResNet are not significant. These five approaches are clustered in the best group. Analogously, the rest approaches are grouped into two clusters based on the T-test results of the MPCE scores (Figure \ref{fig:Ranking}). 

In the best group, BOSS and COTE are all ensemble based models. MCNN exploit convolutional networks but requires heavy preprocessing in data transformation, downsampling and window slicing. Our proposed FCN and ResNet are able to classify time series from scratch and achieves the premium performance. Compared to FCN, ResNet tends to overfit the data much easier, but is still clustered in the first group without significant difference to other four best models. We also note that  the proposed three-layer MLP achieves comparable results to 1NN-DTW without significant difference. Recent advances on ReLU and dropout work quite well in our experiments to help the MLP gain the similar performance with the previous baseline.  

\section{Localize the Contributing Regions with Class Activation Map}
\begin{figure*}[t]
	\centering
	\includegraphics[width=1\textwidth]{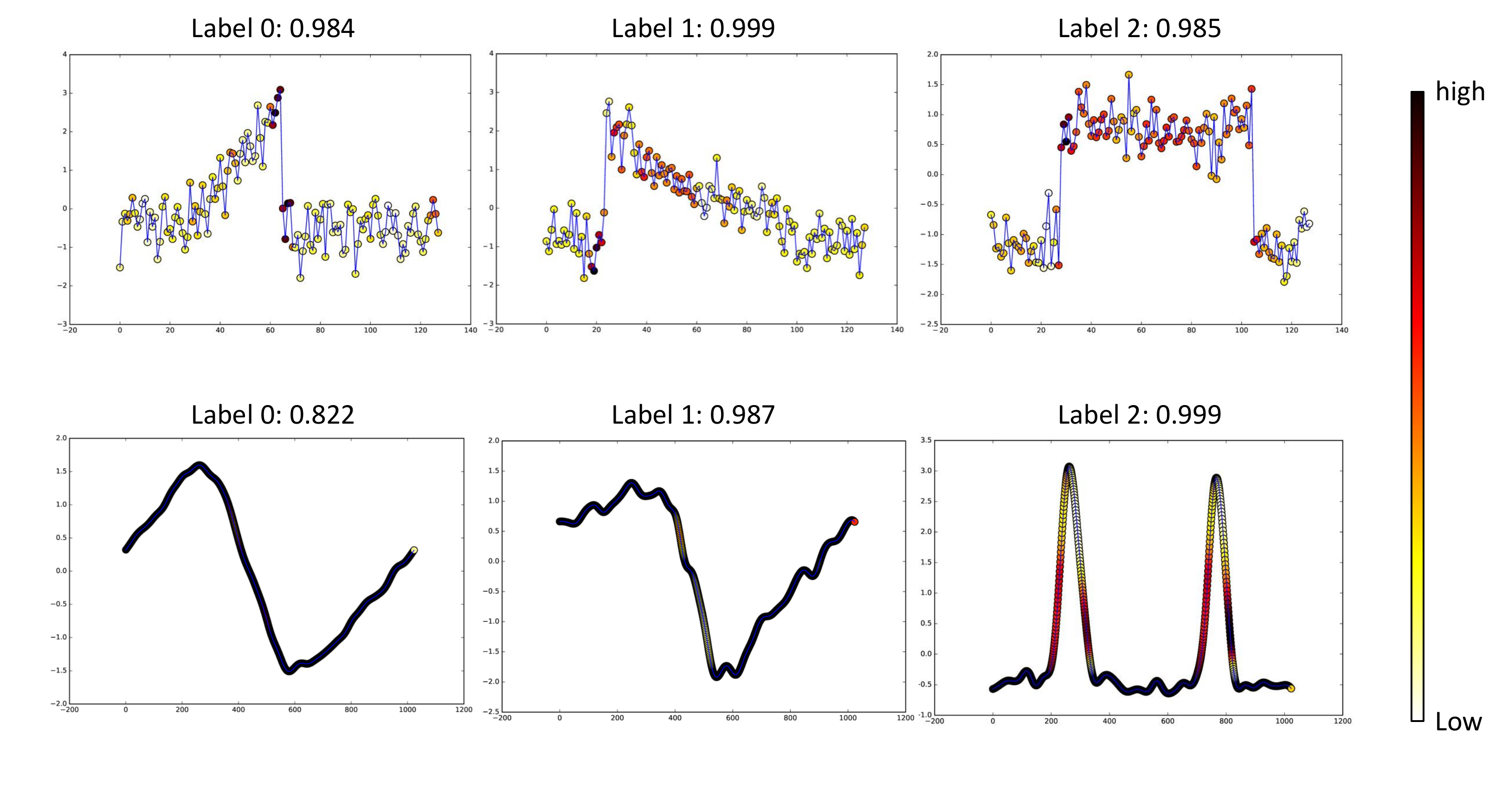}
	\caption{The class activation mapping (CAM) 
		technique allows the classification-trained FCN to both classify
		the time series and localize class-specific regions in a single
		forward-pass. The plots give examples of the contributing regions of the ground truth label in the raw data on the 'CBF' (above) and 'StarLightCurve' (below) dataset.  The number indicates the likelihood of the corresponding label. }
	\label{fig:CAM}
\end{figure*}

Another benefit of FCN with the global average pooling layer is its natural extension, the class activation map (CAM) to interpret the class-specific region in the data \cite{zhou2015learning}.     

For a given time series, let $S_k(x)$ represent the activation of filter $k$ in the last convolutional layer at temporal location $x$. For filter $k$, the output of the following global average pooling layer is $f_k = \sum_{x} S_k(x)$. Let $w_k^c$ indicate the weight of the final softmax function for the output from filter $k$ and the class $c$, then the input of the final softmax function is

\begin{eqnarray}
g_c &=& \sum_{k}w_k^c \sum_{x} S_k(x) \nonumber \\
&=& \sum_{k} \sum_{x} w_k^c S_k(x) \nonumber
\end{eqnarray}

We can define $M_c$ as the class activation map for class $c$, where
each temporal element is given by

\begin{eqnarray}
M_c = \sum_{k} w_k^c S_k(x) \nonumber
\end{eqnarray}

Hence $M_c(x, y)$ directly indicates the importance of the activation at temporal location
$x_i$ leading to the classification of a sequence of time series  to class c. If the output of the last convolutional layer is not the same as the input,  we can still identify the
contributing regions most relevant to the particular category by simply upsampling the class activation
map to the length of the input time series.

In Figure \ref{fig:CAM}, we show two examples of the CAMs output using the above approach.  We can see that the discriminative regions of the time series for the right classes are highlighted. We also highlight the differences in the CAMs for the different labels. The contributing regions for different categories are different. 

On the 'CBF' dataset, label 0 is determined mostly by the region where the sharp drop occurs. Sequences with label 1 have the signature pattern of a sharp rise followed by a smoothly down trending. For label 2, the neural network is address more attention on the long plateau occurs around the middle. The similar analysis is also applied to the contributing region on the 'StarLightCurve' dataset. However, the label 0 and label 1 are quite similar in shapes, so the contributing map of label 1 focus less on the smooth trends of drop down while label 0 attract the uniform attention as the signal is much smoother.

The CAM provides a natural way to find out the contributing region in the raw data for the specific labels. This enables classification-trained convolutional networks to learn to localize without any extra effort. Class activation maps also allow us to visualize the predicted class scores on any given time series, highlighting the discriminative subsequences detected by the convolutional networks. CAM also provide a way to find a possible explanation on how the convolutional networks work for the setting of classification.

\section{Discussion}
\subsection{Overfitting and Generalization}
Neural networks is a strong universal approximator which is known to overfit easily due to the large number of parameters. In our experiments, the overfitting was expected to be significant since the UCR time series data is small and we have no validation/test settings, only choose the model with the lowest training loss for test.

However, our  models generalize quite well given that the training accuracy are almost all 100\%. Dropout improves the generalization capability of MLP by a large margin. For the family of convolutional networks, batch normalization is known to help improve both the training speed and generalization. Another important reason is we replace the fully-connected layer by the global average pooling layer before the softmax layer, which greatly reduces the amount of parameters. Thus, starting with the basic network structures without any data transformation and ensemble, our three models provide very simple but strong baseline for time series classification with the state-of-the-art performance. 

Another nuance of our results is that, deep neural networks work potentially quite well on small dataset as we expand their generalization by recent advances in the network structures and other technical tricks.

\subsection{Feature Visualization and Analysis}
We adopt the Gramian Angular Summation Field (GASF) \cite{wang2015imaging} to visualize the filters/weights in the neural networks. Given a series $X = \{x_1, x_2, ..., x_n\}$, we rescale $X$ so that all values fall in the interval
 $[0,1]$ 
\begin{eqnarray}
\tilde{x}_{0}^i = \frac{x_i-min(X)}{max(X)-min(X)}
\label{eqn:rescale0}
\end{eqnarray}

Then we can easily exploit the angular perspective by considering
the trigonometric summation between each point to identify the
correlation within different time intervals. The GASF are defined as
 
\begin{eqnarray}
G   
&=& \begin{bmatrix}
\cos(\phi_i+\phi_j)
\end{bmatrix} 
\\
&=& \tilde{X}' \cdot \tilde{X} - \sqrt{I-\tilde{X}^2}' \cdot \sqrt{I-\tilde{X}^2} 
\label{eqn:GASF} 
\end{eqnarray}

$I$ is the unit row vector $[1,1,...,1]$. By defining the inner product $<x,y> = x\cdot
y-\sqrt{1-x^2} \cdot \sqrt{1-y^2}$ and $<x,y> = \sqrt{1-x^2} \cdot
y- x \cdot \sqrt{1-y^2}$,GASF are actually quasi-Gramian matrices $[<\tilde{x_1},\tilde{x_1}>]$.

We choose GASF because it provides an intuitive way to interpret the multi-scale correlation in 1-D space. $G_{(i,j||i-j|=k)}$ encodes the cosine summation over the points with the striding step $k$ . The main diagonal $G_{i,i}$ is the special case when $k=0$ which contains the original values. 

\begin{figure*}[t]
	\centering
	\includegraphics[width=1\textwidth]{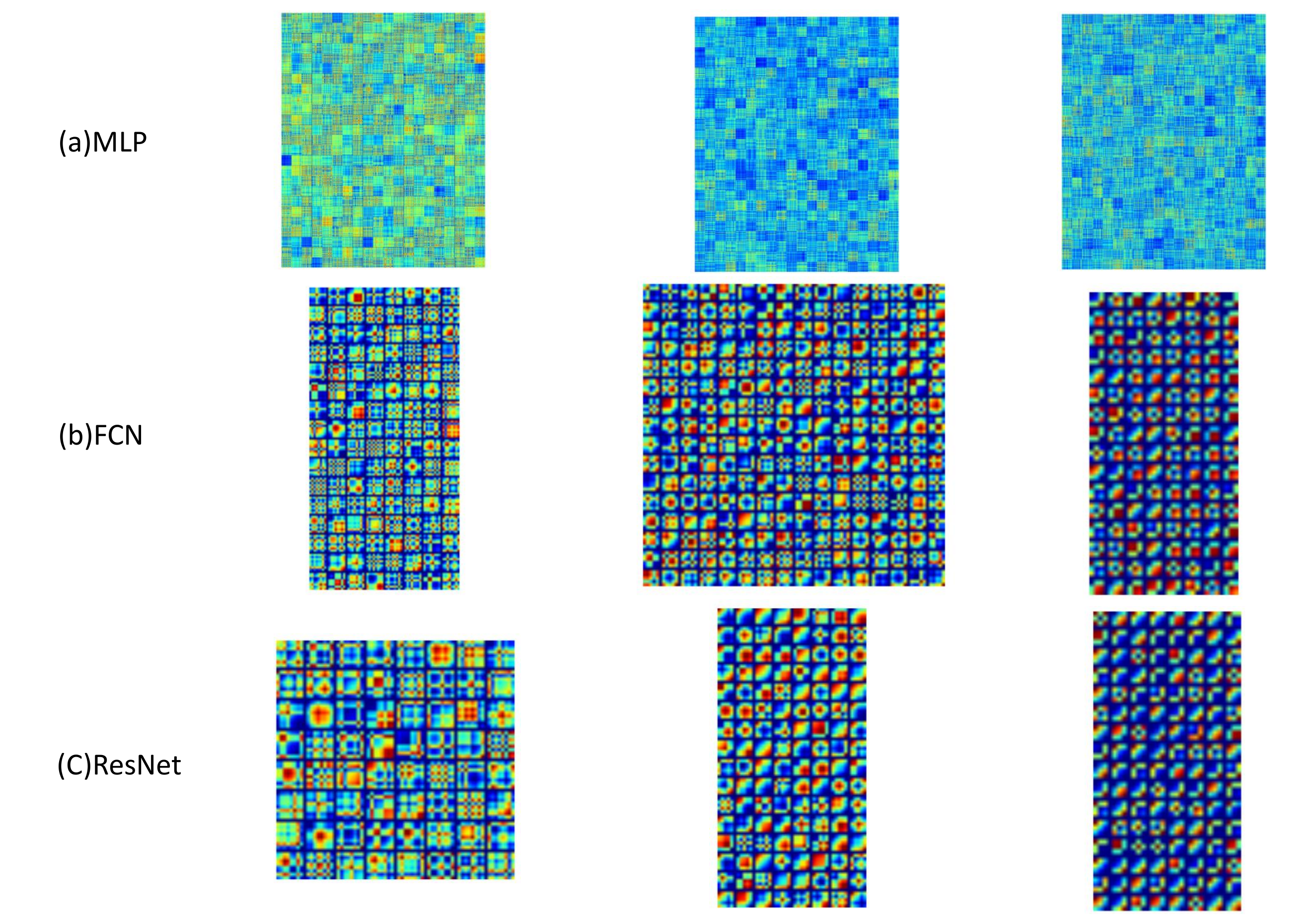}
	\caption{Visualization of the filters learned in MLP, FCN and ResNet on the Adiac dataset. For ResNet, the three visualized filters are from the first, second and third convolution layers in each residual blocks.}
	\label{fig:Feature}
\end{figure*}

Figure \ref{fig:Feature} provides a visual demonstration of the filters in three tested models. The weights from the second and the last layer in MLP are very similar with clear structures and very little degradation occurring. The weights in the first layer, generally, have the higher values than the following layers. 

The filters in FCN and ResNet are very similar. The convolution extracts the local features in the temporal axis, essentially like a weighted moving average that enhances several receptive fields with the nonlinear transformations by the ReLU. The sliding filters consider the dependencies among different time intervals and frequencies. The filters learned in the deeper layers are similar with their preceding layers. This suggests the local patterns across multiple convolutional layers are seemingly homogeneous. Both the visualization and classification performance indicates the effectiveness of the 1-D convolution. 

\subsection{Deep and Shallow}
The exploration on the very deep architecture is interesting and informative. The ResNet model has 11 layers but still holds the premium performance. There are two factors that impact the performance of the ResNet. With shortcut connections, the gradients can flow directly through the bottom layers in the ResNet, which largely improve the interpretability of the model to learn some highly complex patterns in the data. Meanwhile, the much deeper models tend to overfit much easier, requiring more effort in regularizing the model to improve its generalization ability. 

In our experiments, the batch normalization and global average pooling have largely improved the performance in test data but still tend to overfit, as the patterns in the UCR dataset are comparably not so complex to catch. As a result, the test performance of the ResNet is not as good as FCN. When the data is larger and more complex, we encourage the exploration of the ResNet structure since it is more likely to find a good trade-off between the strong interpretability and generalization.

\subsection{Classification Semantics}
\begin{figure*}[t]
	\centering
	\includegraphics[width=1\textwidth]{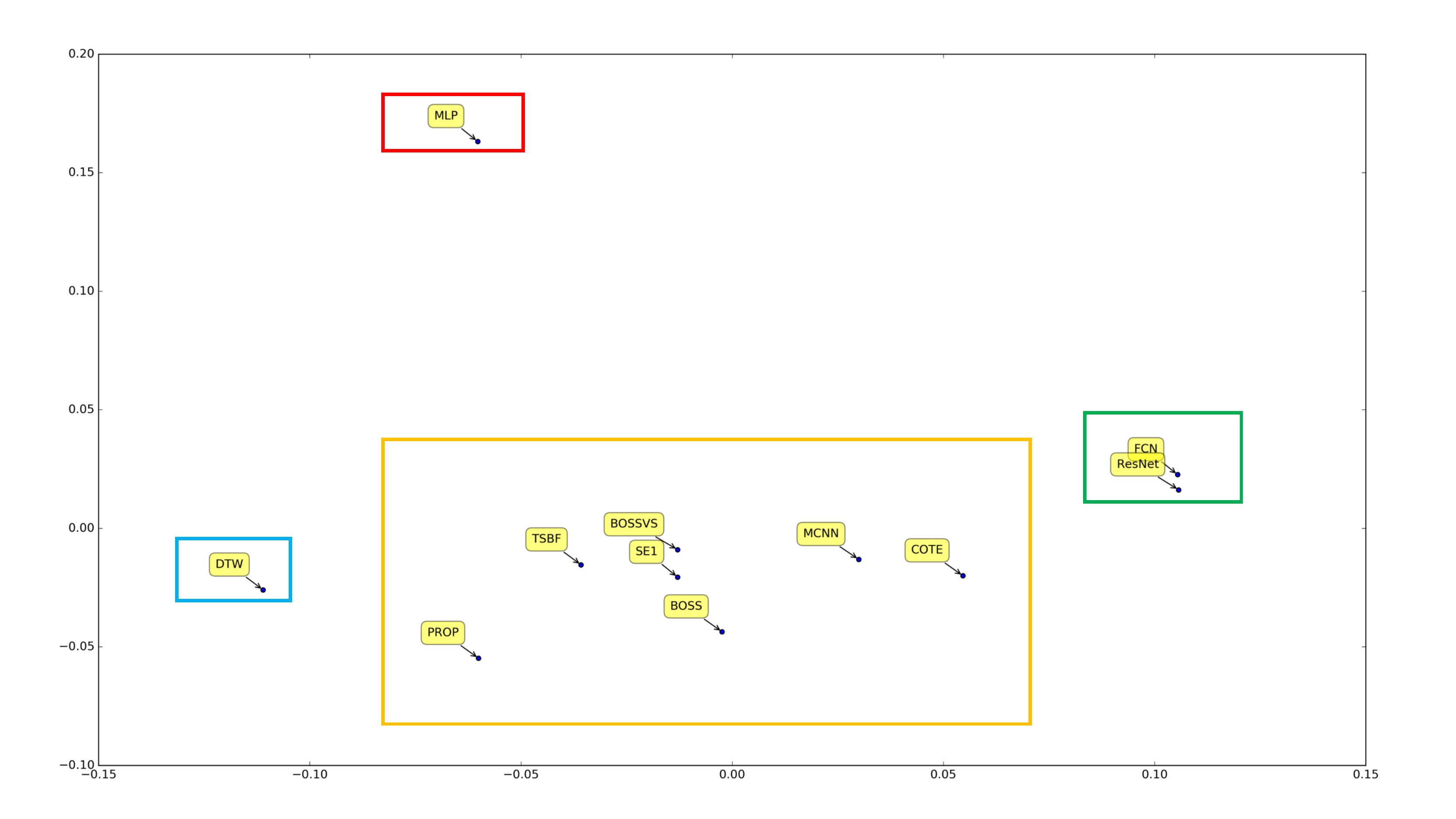}
	\caption{The PCE distribution of different approaches after dimension reduction through PCA.}
	\label{fig:PCA}
\end{figure*}

The benchmark approaches for time series classification could be categorized into three groups: distance based, feature based and neural neural network based. The combination of distance and feature based approaches are also commonly explored to improve the performance. We are curious about the classification behavior of different models as if they all perform similarly on the same dataset, or their feature space and learned classifier are diverged. 

The semantics of different models are evaluated based on their PCE scores. We choose PCA to reduce the dimension because this simple linear transformation is able to preserves large pairwise distances. In Figure \ref{fig:PCA}, the distance between three baseline models with other benchmarks are comparatively large. which indicates the feature and classification criterion learned  in our models are good complement to other models. 

It is natural to see that FCN and ResNet are quite close with each other. The embedding of MLP is isolated into a single category, meaning its classification behavior is quite different with other approaches. This inspires us that a synthesis of the feature learned by MLP and convolutional networks through a deep-and-wide model \cite{cheng2016wide} might also improve the performance. 

\section{Conclusions}
We provide a simple and strong baseline for time series classification from scratch with deep neural networks. Our proposed baseline models are pure end-to-end without any heavy preprocessing on the raw data or feature crafting. The FCN achieves premium performance to other state-of-the-art approaches. Our exploration on the much deeper neural networks with the ResNet structure also gets competitive performance under the same experiment settings. The global average pooling in our convolutional model enables the exploitation of the Class Activation Map (CAM) to find out the contributing region in the raw data for the specific labels. A simple MLP is found to be identical to the 1NN-DTW as the previous golden baseline. An overall analysis is provided to discuss the generalization of our models, learned features, network structures and the classification semantics. Rather than ranking based criterion, MPCE is proposed as an unbiased measurement to evaluate the performance of multiple models on multiple datasets. Many research focus on time series classification and recent effort is more and more lying on the deep learning approach for the related tasks. Our baseline, with simple protocol and small complexity for building and deploying,  provides a default choice for the real world application and a good starting point for the future research.

\begin{table}[h]
	\centering
	\caption{Appendix: The p-values of Wilcoxon rank-sum test between our baseline models with other approaches. }
	\begin{tabular}{rrrr}
		\toprule
		& MLP   & FCN   & ResNet \\
		\midrule
		DTW   & 0.7575 & \textbf{0.0203} & \textbf{0.0245} \\
		COTE  & \textbf{0.0040} & 0.8445 & 0.8347 \\
		MCNN  & \textbf{0.0049} & 0.9834 & 0.9468 \\
		BOSSVS & 0.1385 & 0.1660 & 0.1887 \\
		PROP  & 0.0616 & 0.2529 & 0.2360 \\
		BOSS  & \textbf{0.0076} & 0.8905 & 0.8740 \\
		SE1   & 0.1299 & 0.0604 & 0.0576 \\
		TSBF  & 0.1634 & 0.0715 & 0.0811 \\
		MLP   &  /\     & \textbf{0.0051} & \textbf{0.0049} \\
		FCN   & \textbf{0.0051} &   /\    & 0.9169 \\
		ResNet & \textbf{0.0049} & 0.9169 &  /\ \\
		
		\bottomrule
	\end{tabular}%
	\label{tab:Mann–Whitney}%
\end{table}%

\begin{table*}[t]
	\centering
	\caption{Appendix: The p-values of the paired T-test of the means for the MPCE score on  11 benchmark models. }
	\begin{tabular}{rrrrrrrrrrrr}
		\toprule
		& DTW   & COTE  & MCNN  & BOSSVS & PROP  & BOSS  & SE1   & TSBF  & MLP   & FCN   & ResNet \\
		\midrule
    DTW   &       & \textbf{2.056E-05} & \textbf{5.699E-05} & 5.141E-02 & \textbf{4.832E-05} & \textbf{2.760E-04} & \textbf{3.040E-03} & \textbf{1.311E-02} & 4.234E-01 & \textbf{1.451E-04} & \textbf{3.427E-04} \\
    COTE  &       &       & 2.287E-01 & \textbf{3.721E-05} & \textbf{5.911E-03} & 1.033E-01 & \textbf{1.208E-04} & \textbf{3.528E-04} & \textbf{5.240E-05} & 3.978E-01 & 4.351E-01 \\
    MCNN  &       &       &       & \textbf{3.652E-04} & \textbf{1.354E-02} & 2.497E-01 & \textbf{3.634E-03} & \textbf{3.360E-03} & \textbf{8.023E-05} & 2.495E-01 & 3.757E-01 \\
    BOSSVS &       &       &       &       & 2.140E-01 & \textbf{6.404E-04} & 1.763E-01 & 4.335E-01 & \textbf{4.628E-02} & \textbf{2.983E-03} & \textbf{5.067E-03} \\
    PROP  &       &       &       &       &       & \textbf{3.739E-02} & 4.654E-01 & \textbf{1.440E-01} & \textbf{2.061E-02} & \textbf{2.673E-02} & \textbf{4.241E-02} \\
    BOSS  &       &       &       &       &       &       & \textbf{2.871E-02} & \textbf{1.759E-02} & \textbf{1.049E-03} & 1.879E-01 & 2.751E-01 \\
    SE1   &       &       &       &       &       &       &       & 1.770E-01 & \textbf{9.901E-03} & \textbf{1.208E-02} & \textbf{3.251E-02} \\
    TSBF  &       &       &       &       &       &       &       &       & 7.088E-02 & \textbf{1.510E-03} & \textbf{1.640E-03} \\
    MLP   &       &       &       &       &       &       &       &       &       & \textbf{6.832E-05} & \textbf{3.045E-04} \\
    FCN   &       &       &       &       &       &       &       &       &       &       & 2.508E-01 \\
    ResNet &       &       &       &       &       &       &       &       &       &       &  \\
    \bottomrule
	\end{tabular}%
	\label{tab:PCE-test}%
\end{table*}%




%
\bibliographystyle{IEEEtran}
\bibliography{IEEEabrv,IEEEexample}

\end{document}